\documentclass{ifacconf}
\usepackage{fancyhdr}
\usepackage{color} 
\usepackage{algorithm,algorithmicx,algpseudocode}
\usepackage{graphicx}      % include this line if your document contains figures
\usepackage{natbib}        % required for bibliography
 
\usepackage{tabularht,tabularkv,tabularx}
\usepackage{amsmath,amssymb,amsfonts}
\usepackage{multirow}
\usepackage{bm}
\usepackage{subfigure}
\usepackage{mathtools}
\usepackage{threeparttable}
\usepackage{diagbox}
\usepackage{pifont}% http://ctan.org/pkg/pifont
\newcommand{\cmark}{\ding{51}}%
\newcommand{\xmark}{\ding{55}}%

\newtheorem{definition}{\textbf{Definition}}
\begin{document}
\begin{frontmatter}
	\title{A Hierarchical Collision Avoidance Architecture for Multiple Fixed-Wing UAVs in an Integrated Airspace\thanksref{footnoteinfo}}

\thanks[footnoteinfo]{This research was supported by National Natural Science Foundation of China under No. 61973309.}

\author[First]{Yajing Wang} 
\author[First]{Xiangke Wang} 
\author[First]{Shulong Zhao}
%\author[Second]{Changbin Yu}
\author[First]{Lincheng Shen}
\address[First]{National University of Defense Technology, 
   Changsha, P.R.China \\(e-mail: Wangyajing12@nudt.edu.cn, xkwang@nudt.edu.cn, jaymaths@nudt.edu.cn, lcshen@nudt.edu.cn).}
%\address[Second]{Westlake University, 
%  Hangzhou, P.R.China \\(e-mail: Brad.Yu@anu.edu.au)}

\begin{abstract}                % Abstract of not more than 250 words.
%The inherent nonlinear dynamics and motion constraints, as well as the increasing scale of multiple unmanned aerial vehicles (UAVs) in confined environments, make the safety management of the whole system a very complex and challenging issue. Moreover, the perceptual inaccuracy and failures as well as communication delay and interrupts further enhance the difficulty of solving this problem. 
%
%Notably, the aim of safety management of a multi-UAV system is to avoid colliding with both environmental obstacles and neighbour aircrafts, which is a key issue for the autonomy of multi-UAV systems. 
%
This paper studies the collision avoidance problem for autonomous multiple fixed-wing UAVs in the complex integrated airspace. 
By studying and combining the online path planning method, the distributed model predictive control algorithm, and the geometric reactive control approach, a three-layered collision avoidance system integrating conflict detection and resolution procedures is developed for multiple fixed-wing UAVs modeled by unicycle kinematics subject to input constraints. 
The effectiveness of the proposed methodology is evaluated and validated via test results of comparative simulations under both deterministic and probabilistic sensing conditions.
\end{abstract}

\begin{keyword}
Multiple fixed-wing UAVs, conflict detection and resolution, collision avoidance, hierarchical architecture.
\end{keyword}

\end{frontmatter} \thispagestyle{fancy}
%===============================================================================

%% There are a number of predefined theorem-like environments in
%% ifacconf.cls:
%%
%% \begin{thm} ... \end{thm}            % Theorem
%% \begin{lem} ... \end{lem}            % Lemma
%% \begin{claim} ... \end{claim}        % Claim
%% \begin{conj} ... \end{conj}          % Conjecture
%% \begin{cor} ... \end{cor}            % Corollary
%% \begin{fact} ... \end{fact}          % Fact
%% \begin{hypo} ... \end{hypo}          % Hypothesis
%% \begin{prop} ... \end{prop}          % Proposition
%% \begin{crit} ... \end{crit}          % Criterion
\section{Introduction}

\begin{figure}[t]
	\begin{center}
		\includegraphics[width=8.4cm]{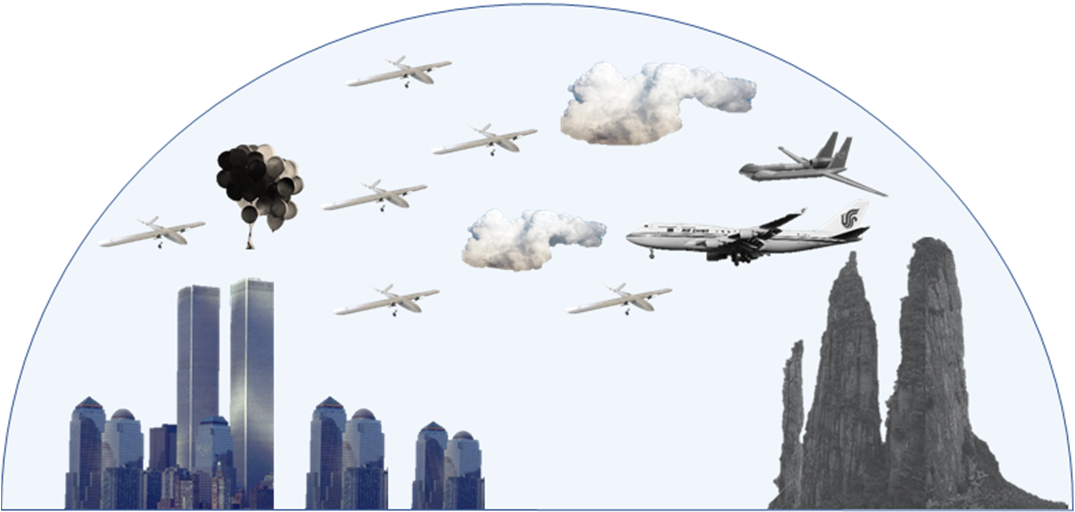}    % The printed column width is 8.4 cm.
		\caption{Prospective mission airspace and possible conflict objects therein} 
		\label{fig:environment}
	\end{center}
\end{figure}

Multiple unmanned aerial vehicles (UAVs) have attracted considerable interest these years, of which prospective applications include disaster area or maritime surveillance, border patrol, environmental sensing, delivery service, etc (\cite{jenie2016taxonomy}). This determines that the UAVs would fly in an integrated airspace with a variety of possible conflict objects therein (See in Fig. \ref{fig:environment}). 
However, one key issue that limits the extensive application and the integration into such complex dynamic integrated airspace system of the UAVs is the collision avoidance problem (\cite{dalamagkidis2008unmanned, dalamagkidis2011integrating, shively2018unmanned}), which is also called as conflict detection and resolution in the literature .
%

%Particularly, multiple fixed-wing UAVs have shown much promising future due to its high speed, long duration, and large load capacity. 

%
Various approaches for collision avoidance of UAVs have been developed these years.
%
%Most of these approaches are proposed for simplex or specific conflict scenarios (\cite{garcia2016biologically},\cite{dentler2019collision} ).
%Several latest review papers have provide discussions about available approaches from different prospectives. For example,
\cite{898217} presented cohesive discussion and comparative evaluation of 68 modeling methods for conflict detection and resolution.  
\cite{lalish2012distributed} discussed the related approaches based on the degree of centralization, the type of the vehicle model, the number of vehicles, and the heterogeneity or homogeneity of the vehicles, respectively. \cite{hoy2015algorithms} mainly reviewed the development of model predictive control (MPC), sensor-based boundary following, sensor-based path planning, and some reactive methods on collision avoidance. Moving obstacles and multi-vehicle situations were also discussed.  \cite{mahjri2015review} summarized the functions of a collision avoidance system into three steps: the sensing, the detection, and the resolution, and reviewed the related approaches from these three aspects. Besides, \cite{zhang2018survey} presented an overview of collision avoidance approaches in large, middle and small scales, respectively.

The above mentioned survey papers summarized the related research from many different aspects. But one common fact indicated by these papers is that most of these approaches are designed for some specific conflict scenarios (\cite{garcia2016biologically},\cite{dentler2019collision} ). This means any single approach cannot be used to completely solve the problem.

% (See table~\ref{tb:Algorithm_comparation}). 
To study out a solution for general conflict resolution in the complex dynamic integrated airspace, \cite{jenie2016taxonomy} firstly proposed a taxonomy of conflict detection and resolution approaches for UAVs based on their types of surveillance, coordination, maneuver, and autonomy, then discussed possible combinations of available approaches for a complete solution. However, specific implementations of such approach combinations were not given.

Therefore, this paper aims to design a hierarchical collision avoidance system, which is capable of detecting and resolving general conflicts, for autonomous multiple fixed-wing UAVs in  the complex dynamic integrated airspace. 

The main contribution of this paper is the proposal and implementation of the hierarchical collision avoidance architecture.
\begin{itemize}
	\item[-] Firstly, a three-layered collision avoidance architecture dependent on local communication and onboard sensing is proposed for multiple fixed-wing UAVs, by analyzing characteristics of existing methods and hierarchical modeling of the local airspace.
	
	\item[-] Then a specific algorithm implementation is studied for each layer of the collision avoidance architecture.
	
	\item[-]Finally, the effectiveness of the proposed methodology is evaluated and validated by comparative simulations carried out under both deterministic and probabilistic sensing conditions.

\end{itemize}
 % It integrates an outer-layer path-planning scheme, a middle-layer distributed model predictive control (DMPC) approach, and an inner-layer geometric reactive method. Statistical results of simulation tests are provided to evaluate and verify the effectiveness of the proposed methodology.

\section{Problem Formulation}

\subsection{Preliminary concept definition}

Before further discussion, two concepts should be clarified: 
\begin{definition}[\textbf{Collision}]\label{def_1}
	For the $i$-th UAV in a $n$-UAV system ($i\in\{1,\cdots,n\}$) and any possible conflict object $o$ in the airspace, a collision happens if
	\begin{equation}\label{eq:collision_condition}
	d_{i,o}\leq R_s
	\end{equation}
	where $d_{i,o}$ represents the distance between the $i$-th UAV and the conflict object $o$, $R_s$ denotes the restricted safe radius of the UAVs. 
	
\end{definition}

\begin{definition} [\textbf{Conflict}]
	For a UAV, a conflict is detected if a collision is predicted to happen on it within a specific time period $\tau_w$ in the future, where $\tau_w$ is the early warning time for collision conflicts.
\end{definition}

Then two main  functions of collision avoidance control are to firstly detect potential conflicts and then take actions to avoid collisions if any conflicts are detected.
% is to firstly detect and then resolve potential collision conflicts. The resolution step is actually a procedure of implementing collision avoidance  maneuvers. So the primary basis for evaluating the effectiveness of a conflict resolution scheme is whether or not the collision is successfully avoided in the end.

\subsection{Conflict scenarios analysis}
A collision avoidance system aims to enable the UAVs to handle all possible collision conflicts to ensure safe and orderly operations. To this end, various possible conflict objects in the complex integrated airspace are first discussed. See Table \ref{tb:obstacle_categorization}.
\begin{table}[htb]
	\caption{Classifications of various conflict objects in the integrated airspace}\label{tb:obstacle_categorization}	
	\begin{center}	
		\renewcommand{\arraystretch}{1.2}
		\begin{threeparttable}
			\begin{tabular}{c|c|c|c}
				\hline
				\multicolumn{2}{c|}{ Classification Principles}  &  Static    &Dynamic  \\\hline
				\multirow{6}{*}{Non-cooperative} & 	\multirow{3}{*}{Unknown}  & \multirow{3}{*}{new buildings} & birds \\ 
				~                                           & ~                                    & ~                                         & air masses \\
				~                                       & ~                                    & ~                                         & enemy UAVs \\ \cline{2-4}
				~                                           & 	\multirow{3}{*}{Known}    & mountains                              & ~                 \\
				~                                           &   ~                                  &  old buildings                         & ~\\
				~                                        &   ~                                &  lighthouses                          & ~      \\\hline
				\multirow{3}{*}{Cooperative}        &\multirow{2}{*}{Unknown}    & ~                                         &civil aircrafts\\
				~                                        & ~                                    & ~                                         & other UAVs \\ \cline{2-4}
				~                                           & Known                               & ~                                        & neighbor UAVs \\ \hline
			\end{tabular}
		\end{threeparttable}
	\end{center}
\end{table}

 Firstly, in the consideration of motion states, conflict objects are classified as static and dynamic. 
 Then according to whether there is active avoidance intention in the process of conflict resolution, they are classified into cooperative and non-cooperative ones. For example, objects like flying birds, balloons, and air masses,  which are very much likely to disturb the flight but cannot implement active avoidance if conflicts exist, are classified as non-cooperative.
  Civil aircraft are treated as cooperative because generally they can take active collision avoidance maneuvers based on some common rules, although the unknown nature of UAVs to the civil aircraft and vice versa make the cooperation rather challenging. 
  Thirdly, based on the ways of information acquisition, those obtained by prior knowledge or active communications are included in the known category. Other objects like some new buildings or other aircraft, requiring real-time perception, are included in the unknown category.

% Timely and accurate collection of spatial status information is a preliminary requisite for effective safety management. For a given task area, most information about static objects, such as buildings and mountains, can be obtained from the satellite map. Only when the map is unavailable or some new buildings exist in this area, it would depend on the onboard sensing system to gain such information. Situations of dynamic objects could be more diverse. Flying birds, balloons, and air masses,  which are very much likely to disturb the flight and cannot implement active avoidance if conflicts exist. Civil aircrafts usually can take active collision avoidance maneuvers. However, the unknown nature of UAVs to the civil aircrafts and vice versa make the success of collision avoidance rather challenging. Moreover, intersections of routes or other unexpected events can cause conflicts between multiple UAVs themselves. The difference is, that the UAVs can communicate with each other and thus can implement a coordinated conflict resolution strategy.

\subsection{Collision avoidance objective}

This paper mainly studies real-time online collision avoidance. Therefore, those known environmental objects,  that can generally be handled before the flight through trajectory pre-planning, are not the focus of this paper.  For the rest of the conflict objects, taking the $i$-th UAV in a $n$-UAV system as a reference, denote the set of its neighbor UAVs as $\mathcal{N}_i$, the set of other potential unknown conflict objects as $\mathcal{O}_i$. Then all possible conflict objects of the $i$-th UAV can be represented as the augmented obstacle set:
$$\mathcal{O}_i^{aug}\coloneqq \mathcal{N}_i\cup\mathcal{O}_i$$
Then according to Definition \ref{def_1}, the primary objective of collision avoidance control would be to keep a separate distance  larger than $R_s$ for the $i$-th UAV from all obstacles in $\mathcal{O}_i^{aug}$, e.g., to ensure
\begin{equation}\label{eq:CA_requirement}
d_{i,o} > R_s, \forall o\in\mathcal{O}_i^{aug}
\end{equation}

%Moreover, fixed-wing UAVs are subject to dynamic constraints of the minimum cruising speed. The limited load capacity and duration, and specific task performance index should also be considered.

Moreover, 
except for the collision avoidance requirement in \eqref{eq:CA_requirement}, dynamic constraints of the minimum cruising speed and limited heading rate,  and optimization for the maneuver energy consumption and the required task performance index should also be considered in the collision avoidance strategy.

\subsection{Kinematics}

This paper studies the collision avoidance problem for UAVs implementing planar flights. Thus the fixed-wing UAVs are modeled as unicycle kinematics:
\begin{equation}\label{eq:kinematics}
\begin{cases}
\dot x =v\cos\phi \\ 
\dot y=v\sin\phi \\ 
\dot\phi = u
\end{cases}
\end{equation}
where, $(x,y,\phi)^T$ represents the state vector of the UAV, $(x,y)^T$ denotes the position and $\phi$ describes the heading angle,  $v$ is the cruising speed, which is set to be constant during the flight, and the control input $u = \omega$ denotes the heading rate of the UAV. 
Meanwhile, the control input is subject to the following constraint:
\begin{equation}
u \in \mathcal{U}, \mathcal{U}\coloneqq\{\omega|-\omega_{max}\leq \omega \leq \omega_{max}\}
\end{equation}
where $\omega_{max}$ represents the upper bound of the heading rate.

Considering the discrete control process during the flight,  we use the second-order Runge-Kutta method to obtain the discrete kinematics model.

\section{Hierarchical collision avoidance architecture}

\newcommand{\tabincell}[2]{\begin{tabular}{@{}#1@{}}#2\end{tabular}} 
\begin{table*}[htb]
	\caption{Algorithm review}\label{tb:Algorithm_comparation}	
	\begin{center}	
		\renewcommand{\arraystretch}{1.2}
		\begin{threeparttable}
			\begin{tabular}{c|c|cc|ccc|c}
				\hline\hline
				~ & ~& 	\tabincell{c}{Computation \\ complexity}& Optimality & MV & MO & IRM  & References \\\hline
				\multirow{4}{*}{\tabincell{c}{Path \\ planning}} & Graph search approaches & high &  $\blacktriangle$ &\cmark&$
				\blacktriangle$&$\blacktriangle$&[1],[2]\\
				~ & Mathematical programming   &  high&  \cmark& \cmark & $\blacktriangle$& \cmark&[1],[5] \\
				~  & Artificial heuristic approaches & high & \cmark  & \cmark&$\blacktriangle$& \cmark&  [3],[5]  \\
				~  & Potential field based planning&  low  & $\blacktriangle$  &\cmark&$\blacktriangle$&$\blacktriangle$&  [1]\\ \hline
				\multirow{2}{*}{\tabincell{c}{Optimized \\ control}} & Game theory based approaches & high & $\blacktriangle$& \cmark& \xmark& \cmark & [6],[7]\\
				~ & Distributed model predictive control & $\bigstar$ & \cmark & \cmark  &\cmark&\cmark& [4] \\ \hline
				\multirow{3}{*}{\tabincell{c}{Reactive \\ approaches}} & Geometric approaches & low & \xmark &$\blacktriangle$&\cmark& $\blacktriangle$ & [4]\\
				~ & Rule-based approaches & low &  \xmark&\cmark&\xmark&$\blacktriangle$&[1]\\
				~ & Potential field based reactive approaches & low & \xmark &\cmark&\cmark&$\blacktriangle$& [1],[4]\\
				\hline
				\hline
			\end{tabular}
			\begin{tablenotes}
				\footnotesize
				\item[*] Reference: [1] \cite{zhang2018survey}, [2] \cite{dadkhah2012survey}, [3] \cite{yu2015sense}, [4] \cite{hoy2015algorithms}, [5] \cite{mahmoudzadeh2018online}, [6] \cite{mylvaganam2018autonomous}, [7] \cite{mylvaganam2017differential}
				\item[*] Key: MV (\textbf{M}ultiple \textbf{V}ehicles), MO (\textbf{M}oving \textbf{O}bstacles), IRM (\textbf{I}nput \textbf{R}estricted \textbf{M}odel)
				\item[*] Symbols: $\bigstar$ (Not necessarily high), $\blacktriangle$ (With some disadvantages).
			\end{tablenotes}
		\end{threeparttable}
	\end{center}
\end{table*}

\subsection{Three-layered collision avoidance framework }

\begin{figure}
	\begin{center}
		\includegraphics[width=6.0cm]{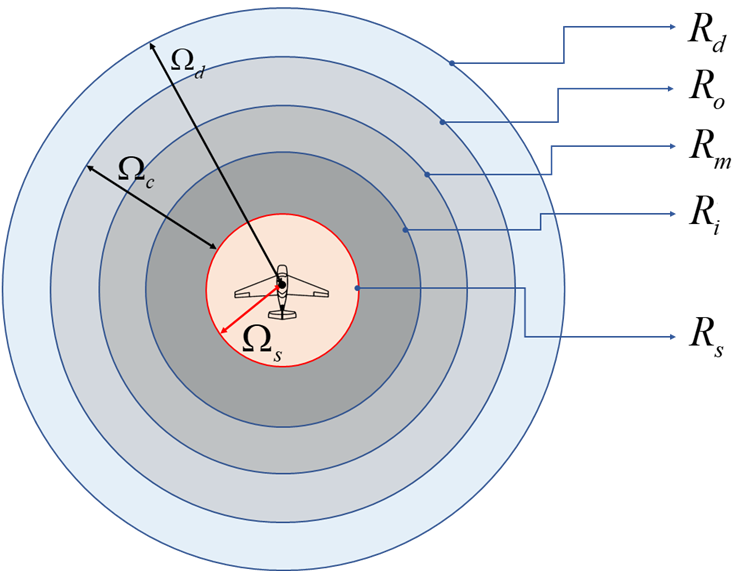}    % The printed column width is 8.4 cm.
		\caption{The three-layered conflict detection region} 
		\label{fig:local_airspace_partition}
	\end{center}
\end{figure}
%The SMSs equipped on the UAVs depend on the onboard hardware senser and communication systems for spacial status information collection, and softwares of specific conflict detection and resolution algorithms for collision avoidance. This paper focuses on the software part to provide a complete distributed solution of conflict detection and resolution for autonomous multiple fixed-wing UAVs suject to all possible conflict obstacles. 

%However, in addition to the diversity of possible conflict obstacles in $\mathcal{O}^{aug}$,  the dynamic situation also varis greatly at different ranges from the UAV, which make it rather challenging for any single algorithm to detect or resolve all possible conflicts. This naturally inspires an idea of combining several effective and complementary algorithms into a complete hierarchical SMS.

The two main functions of collision avoidance control can be briefly described as conflict detection and resolution. Conflict detection using one single approach once for all can easily fail or delay because of sensing inaccuracy and uncertainty, or communication delay and interrupts.
Besides, approaches for conflict resolution in the literature have different advantages and disadvantages in different conflict situations. 
Therefore, a three-layered collision avoidance architecture including a three-layered airspace partition for hierarchical conflict detection and a three-layered complementary conflict resolution strategy is proposed in this subsection.

\subsubsection{\textbf{Three-layered airspace for hierarchical conflict detection}}

Dynamic properties at different ranges from the UAV can vary greatly. 
Thus, a conflict detection region $\Omega_c$ is  introduced and partitioned into three layers to implement hierarchical conflict detection:
\begin{equation}
\begin{split}
	&\qquad\Omega_c=\Omega_o \cup \Omega_m \cup \Omega_i \\
	&\begin{cases}
\Omega_o& = \{\bm{P}| R_m < d_{\bm{P}} \leq R_o\leq R_d\}\\
\Omega_m &=\{\bm{P} | R_i < d_{\bm{P}} \leq R_m\}\\
\Omega_i &= \{\bm{P} | R_s < d_{\bm{P}} \leq R_i \}
\end{cases}
\end{split}
\label{eq:conflict_region_partition}
\end{equation}
where $R_o$, $R_m$ and  $R_i$ are the radius of the three-layered conflict detection airspace, $d_{\bm{P}}$ denotes the distance of point $\bm{P}$ in the nearby airspace from the UAV. See Fig. \ref{fig:local_airspace_partition}. Note that $\Omega_d\supseteq\Omega_c$ is the perceptible area of the UAV.

The outer-layer airspace has quite long distance from the UAV, which indicates that conflict situations in this area are essentially determined to the reference flight trajectories. Situations in middle-layer airspace is the most dynamic and complex. Motion state variations of neighbor UAVs, other aircraft,  balloons, and the UAV itself, increase the uncertainty of conflict situations in this area.  The inner-layer airspace has very short distance from the UAV, which determines the UAV should be able to detect potential conflicts very quickly so as to leave enough time for collision avoidance actions. Therefore, a hierarchical conflict detection and resolution scheme is developed in the consideration of these properties.

\subsubsection{\textbf{Three-layered conflict detection and resolution}}

% As previously discussed,  in addition to the diversity of possible conflict obstacles in $\mathcal{O}^{aug}$,  the dynamic situation also varies greatly at different ranges from the UAV, which makes it rather challenging for any single algorithm to detect or resolve all possible conflicts. This naturally inspires an idea of combining several effective and complementary algorithms into a hierarchical SMS.
 
\begin{figure}[h]
	\centering
	\includegraphics[width=8.4cm]{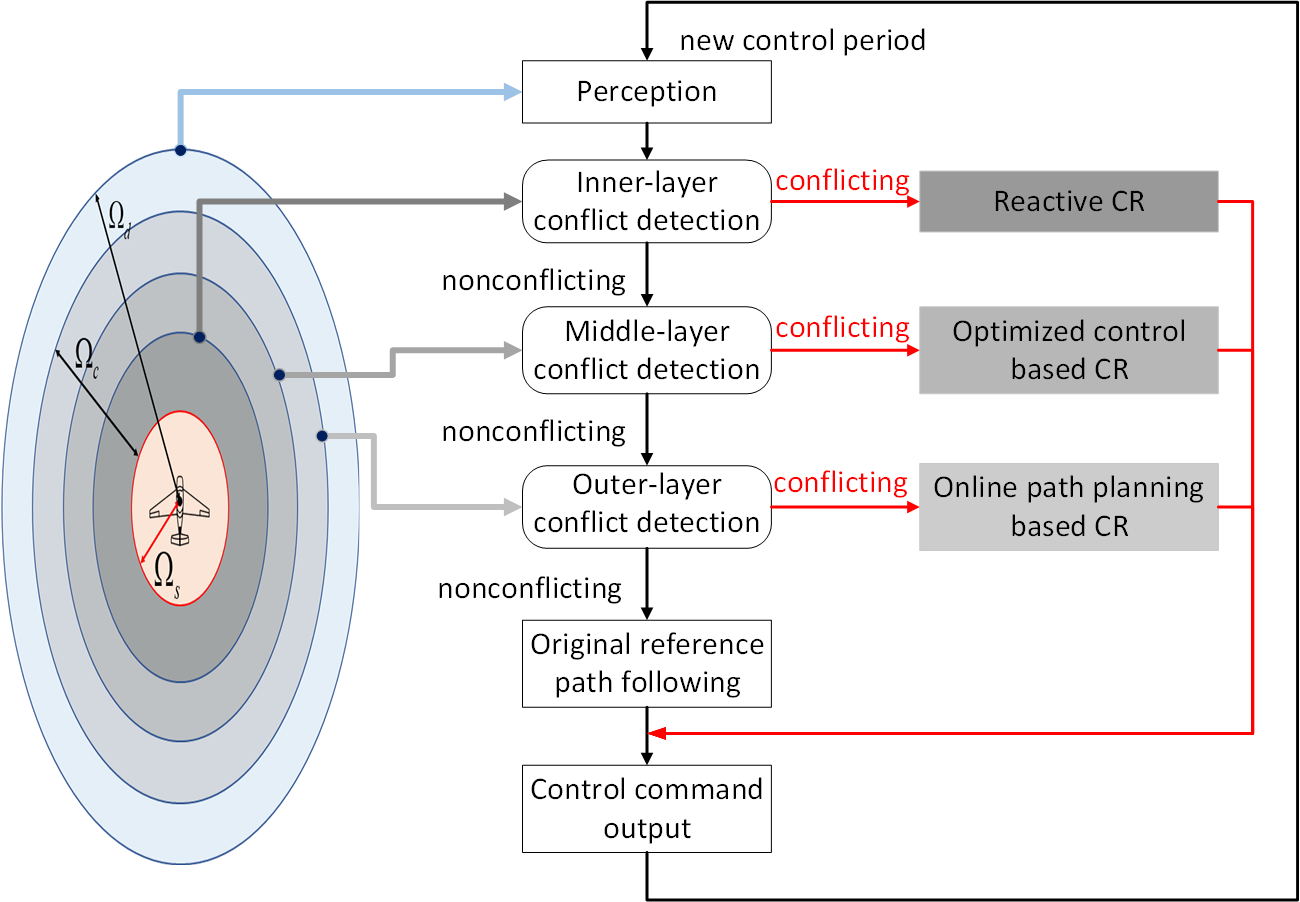}
	\caption{The three-layered collision avoidance framework (CR: the abbreviation of "conflict  resolution")}
	\label{fig:CA_scheme_framewrk}
\end{figure}

Approaches for conflict resolution in the literature can be roughly classified into three categories: path planning, optimized control, and reactive approaches. See Table \ref{tb:Algorithm_comparation}. To maximize the advantages of different algorithms, a hierarchical collision avoidance framework integrating these three types of algorithms is proposed for general conflict scenarios. See Fig. \ref{fig:CA_scheme_framewrk}. 

% The outer-layer airspace gives the UAV relatively long time for avoidance actions, so The path planning scheme is chosen for conflicts in the outer-layer airspace for its smooth  this area. The high uncertainty and dynamics of the middle-layer airspace requires better optimization and flexibility of the conflict detection and resolution approaches, for which a model predictive controller should be very suitable.
%The inner-layer airspace has shortest distance with the UAV, which  leaves the UAV very little time for collision avoidance if any conflict is detected. So the reactive control approach is chosen for this layer of the SMS. 

Considering the range from the UAV and the level of dynamic complexity, a hierarchical collision avoidance framework integrating path planning schemes for the outer layer, optimized control for the middle layer and reactive methods for the inner layer is developed.

Notably, the inner-layer reactive control law has the highest priority when it is activated. The middle-layer optimized control scheme  has the second priority, which can provides better optimization and flexibility for highly dynamic middle-layer airspace. When there is no conflict detected, the UAVs fly according to the scheduled trajectories.

\subsection{Methodology}

%On the basis of the proposed three-layered framework for SMS above, 
This subsection studies to present an implementation for the proposed hierarchical collision avoidance framework.
% which combines an online path planning algorithm, a distributed model predictive controller, and a geometric reactive approach.

\subsubsection{\textbf{Outer-layer path planning using sub-targets and Cubic B-spline}}

 Path planning approaches have been widely studied for collision avoidance problems. \cite{shuai2014real} proposed a real-time obstacle avoidance method using a sub-targets algorithm and Cubic B-spline for mobile robots that move to a specified target point.
Inspired by his work, a conflict detection scheme based on the closest point of environment obstacles from the reference flight path is developed,  with consideration of flight tracking error.
 This approach relies on the onboard sensing system for spacial status information updating.

In this way, the sub-targets generation procedure in \cite{shuai2014real} is extended to curved-path following scenarios. Then a collision-free smooth path is generated using the sub-targets and Cubic B-spline algorithms as in \cite{shuai2014real}.

\subsubsection{\textbf{Middle-layer DMPC-based collision avoidance}}

Distributed model predictive control (DMPC) can explicitly deal with inter-agent constraints and find approximate optimal solutions for subsystems. Besides, the state prediction of MPC provides prior advantage in conflict detection. Thus a DMPC collision avoidance strategy, which executed by all the subsystems synchronously, is developed. The distributed controllers will rely on the local communication system and onboard sensing system for environmental information collection.

Firstly, the conflict detection procedure based on state prediction is implemented. Since the reference trajectory is already known, the reference state of each UAV in the future could be computed and transmitted to its neighbor UAVs with the newest state information. Then for the $i$-th UAV in a $n$-UAV system, the assumed motion states of all neighbor UAVs in $\mathcal{N}_i$ could be computed. Also, the sensing system obtains the real-time information of environmental objects in $\mathcal{O}_i$. Thus the distance variations of the UAV from its neighbor UAVs and other environmental objects, e.g., all obstacles in $\mathcal{O}_i^{aug}$, could be predicted for conflict detection.

Then if any conflict is detected at time interval $k$,  the optimal local collision avoidance input sequence $\bm{u}_{i,(k)}^{\ast} = \{u_{i,(k+0|k)}^{\ast},\cdots,u_{i,(k+N-1|k)}^{\ast}\}$ would be generated  by solving the following optimization problem:
\begin{equation}
\label{eq:DMPC_scheme}
\begin{split}
&J_{i,(k)}^{\ast}=\min_{\bm{u}_{i,(k)}}J_{i,(k)}\left(\bm{X}_{i,(k)},\bm{u}_{i,(k)}, \tilde{\bm{X}}_{(k-1)}^{\mathcal{O}_{i}^{aug}}\right) \\
&s.t. \\  &\quad u_{i,(k+l|k)}\in\mathcal{U}, \forall l = 0,1,\cdots,N-1
\end{split}
\end{equation}
where $\bm{X}_{i,(k)}$ is the newest state, $\tilde{\bm{X}}_{(k-1)}^{\mathcal{O}_{i}^{aug}}$ represents the predicted motion states of $\mathcal{O}_i^{aug}$. Once the local collision avoidance command sequence $\bm{u}_{i,(k)}^{\ast}$ has been generated, the first item $u_{i,(k+0|k)}^{\ast}$ would be applied to the UAV, and the complete sequence would be transmitted to its neighbor UAVs for next conflict detection. The whole process is summarized in Algorithm \ref{alg:DMPC_based_CDR}. 

Due to the limitations of the length of the paper, this algorithm is not rigorous detailed and a complete description and analysis will be given
in our another paper later.
 
\begin{algorithm}[htbp]
	\caption{Middle-layer DMPC-based collision avoidance}
	\label{alg:DMPC_based_CDR}
	\begin{algorithmic}[1]
		\State Parameter initialization: $T$, $N$, $R_m$, $R_s$, etc.
		\State Spacial status information updating: $\mathcal{O}_i^{aug}$
		\State $k \gets k+1$
		\State Conflict detection based on motion prediction
		\Procedure{Conflict Resolution}{}
		\State Calculate $\bm{u}_{i,(k)}^{\ast}$ by solving \eqref{eq:DMPC_scheme}
		\State Apply $u_{i,(k+0|k)}^{\ast}$ to the UAV
		\State Transmit the newest state and the control sequence $\bm{u}_{i,(k)}^{\ast}$ to neighbor UAVs
		\EndProcedure
		\State Returen to step 2		
	\end{algorithmic}
\end{algorithm}

\subsubsection{\textbf{Inner-layer reactive collision avoidance}}

Inner-layer conflict detection and resolution provides the last guarantee for the flight safety of UAVs. Thus for quick response to conflicts, sufficient conditions for non-conflicting flights of any two UAVs in a short distance were derived in previous work (\cite{8996560}), which is utilized for conflict detection. Then a reactive collision avoidance control law is firstly proposed for two-UAV conflict based on the collision-free conditions:
\begin{equation}\label{eq:inner_layer_CA}
u_i=\rho k_{\psi}\left(\frac{1}{2}\arccos{\frac{\bm{v}_{ij}\cdot\bm{P}_{ij}}{|\bm{v}_{ij}||\bm{P}_{ij}|}}-\pi/4\right)
\end{equation}
where, parameter $\rho$ is the sign of turning direction, $k_{\psi}$ in (1/s) is a constant coefficient, which transforms the desired heading change into the desired heading rate, $\bm{v}_{ij}$ and $\bm{P}_{ij}$ are the relative velocity and position vectors of the $i$-th and the $j$-th UAVs, respectively. 

Moreover, the collision avoidance control law in \eqref{eq:inner_layer_CA} was further developed by integrating some additional rules on direction choosing,  for more complicated conflict scenarios which involves more than two UAVs  (\cite{8996560}). 

\subsection{Overall hierarchical algorithm}

Finally, the overall hierarchical implementation of the hierarchical collision avoidance system is developed by integrating the three approaches described above, which is presented in Algorithm \ref{alg:overall_SMS_algorithm}.

\begin{algorithm}[htbp]
	\caption{The distributed hierarchical collision avoidance for multiple UAVs}
	\label{alg:overall_SMS_algorithm}
	\begin{algorithmic}[1]
		\Procedure{Parameter Initialization}{}
		\State Initializa $\omega_{max}$, $T$, $R_o$, $R_m$, $R_i$, $R_s$, and $N$;
		\State $inner\_conflict\_flag \gets 0$
		\State $middle\_conflict\_flag \gets 0$
		\State $outer\_conflict\_flag \gets 0$
		\EndProcedure
		\State Update data for $\mathcal{O}_{i,(k)}^{aug} = \mathcal{N}_{i,(k)} \cup \mathcal{O}_{i,(k)}$
		\State $k \gets k+1$
		\Procedure{Conflict Detection}{}
				\State \textbf{return}  $inner\_conflict\_flag$,              \State $middle\_conflict\_flag$,         and $outer\_conflict\_flag$
		\EndProcedure
			\Procedure{Conflict Resolution}{}
		\If{$inner\_conflict\_flag==1$}
		\State \textbf{Do} reactive cillision avoidance control
		\ElsIf{$middle\_conflict\_flag==1$}
		\State \textbf{Do} DMPC based collision avoidance
		\ElsIf{$outer\_conflict\_flag==1$}
		\State \textbf{Do} path-planning based collision avoidance
		\Else
		\State \textbf{Do} normal trajectory tracking.
		\EndIf
		\EndProcedure
		\State Return to step 7		
	\end{algorithmic}
\end{algorithm}

\section{Simulations}
Comparative simulation tests for the proposed hierarchical collision avoidance system are carried out in comparison with the DMPC-only collision avoidance approach. The DMPC approach is chosen for comparison because it is a typical algorithm which can deal with various dynamic conflict scenarios in the literature.

\subsection{Simulation settings}
  Simulations are performed on Matlab 2018. Each UAV is functioned as a separate running Matlab and uses the UDP protocol for local communication, which is set to be fully connected. The impact of communication delay and failures are ignored.

  The UAVs utilize the kinematics in \eqref{eq:kinematics} and are required to follow several pre-planned closed triangle-like curved paths at a constant cruising speed using the pure pursuit with line-of-sight approach (\cite{sujit2014unmanned}). To increase the frequency of conflicts for simulation verification, each reference path is designed to be intersected with the others. Each circle of the paths is about $1500m$. Besides, several environmental obstacles are distributed on or near the reference paths. Then during simulation flights, the UAVs perform the collision avoidance method, e.g., the hierarchical collision avoidance system or the DMPC-only approach, when certain conflict is detected. Main parameter settings are presented in Table \ref{tab:list_of_parameters}. 
   
  \begin{table}[htbp]
  	\caption{Parameter settings in simulation tests}
  	\begin{center}
  		\renewcommand{\arraystretch}{1.2}
  		\begin{tabular}{p{0.6cm}|p{1.3cm}|p{5.5cm}}
  			\hline
  			\hline
  			~& \textbf{Value}& \textbf{Meaning} \\
  			\hline
  			$V$&  $19 (m/s)$& The cruising speed  \\
  			\hline
  			$\omega_{max}$& $0.6 (rad/s)$& The maximum heading rate  \\
  			\hline
  			$R_o$& $80 (m)$ & The outer-layer detection region radius  \\
  			\hline
  			$R_m$&  $70 (m)$& The middle-layer detection region radius\\
  			\hline
  			$R_i$ & $55 (m)$& The inner-layer detection region radius\\
  			\hline
  		%	$N$& $10 $ & The number of steps of the model predictive controller\\
  		%	\hline
  			$T$&  $0.1 (s)$& The control and sampling period \\
  			\hline
  			$R_s$& $30 (m) $& The restricted safe radius of the UAV \\
  			%$l_{ws}$& 2-5 (m) & The wingspan of the UAV \\
  			\hline
  			\hline
  		\end{tabular}
  		\label{tab:list_of_parameters}
  	\end{center}
  \end{table}
 
\subsection{Simulations with deterministic sensing}

The simulations are firstly carried out for 5 UAVs with deterministic sensing, e.g., the information of obstacles are obtained as far as they enter the perceptible area $\Omega_d$.  Then, the UAVs keep doing conflict detection during the flight, and activate the corresponding conflict resolution methods when certain conflicts are detected.

In each comparative simulation, the initial positions of the UAVs are the same and randomly chosen from the non-conflict points on the reference paths. The operation time is set to be 5000 control cycles. Thus the flight distance of each UAV in a simulation test is about $9500m$. Once the distance of the UAV from obstacles is less than $R_s$, it is marked as a failure of conflict resolution. Then total number of failures is calculated for comparison. 

\begin{table}[htb]
	\caption{Simulations with deterministic sensing}\label{tb:determinisitic_sensing_5_UAVs}	
	\begin{center}	
		\renewcommand{\arraystretch}{1.2}
		\begin{tabular}{c|c|c|c|c}
			\hline\hline
			~&	\multicolumn{2}{c|}{ \textbf{Failure times}} & \multicolumn{2}{c}{ \tabincell{c}{\textbf{Average }  \\ \textbf{collision-free } \\ \textbf{distance (m)}}} \\ \cline{2-5}
			~&  \tabincell{c}{DMPC \\ only} & \tabincell{c}{Hierarchical  \\ CAS}    & \tabincell{c}{DMPC \\ only} & \tabincell{c}{Hierarchical  \\ CAS} \\ \hline
			Test 1 	& 74 &  43 & 128.38 & 220.93 \\
			Test 2& 	69 & 24  & 137.68 & 395.83  \\ 
			Test 3& 	66 & 17  & 143.94  & 558.82 \\ 
			Test 4& 	77 & 28  & 123.38   & 339.28 \\ 
			Test 5& 	56 & 15  & 169.64   & 633.33 \\ \hline
			Summation & 342 & 127 &  \\\hline
				Mean & ~ & ~ & 140.60 & 429.64 \\\hline\hline
		\end{tabular}
		\begin{tablenotes}
		\footnotesize
		\item{*} CAS: the abbreviation of "collision avoidance system"
	\end{tablenotes}
	\end{center}
\end{table}

Table \ref{tb:determinisitic_sensing_5_UAVs} presents the results of 5 comparative simulations. From the content we can see that the total number of conflict resolution failures in flights of about $47500m$ is $127$ using the proposed hierarchical collision avoidance system, which is much less than the result of the DMPC-only method ($342$). Besides, the mean of average collision-free distance using the hierarchical collision avoidance system is $429.64m$, which is much longer than that of the DMPC-only method ($140.60m$).

\subsection{Simulations with probabilistic sensing}

In the consideration of perception uncertainties in reality, simulations are then performed for 5 UAVs with probabilistic sensing, e.g., obstacles are successfully sensed at a increasing probability as the distance from the UAV decreases.

 In simulation tests, the probability of successful perception in the outer-layer conflict detection region is $0.70$, the probability of the middle-layer region is $0.85$, and that of the inner-layer region is set to be $1$. Results of 5 comparative simulations are presented in Table \ref{tb:probabilistic_sensing_5_UAVs}.

\begin{table}[htb]
	\caption{Simulations with probabilistic sensing}\label{tb:probabilistic_sensing_5_UAVs}	
	\begin{center}	
		\renewcommand{\arraystretch}{1.2}
		\begin{tabular}{c|c|c|c|c}
			\hline\hline
			~&	\multicolumn{2}{c|}{ \textbf{Failure times}} & \multicolumn{2}{c}{ \tabincell{c}{\textbf{Average }  \\ \textbf{collision-free } \\ \textbf{distance (m)}}} \\ \cline{2-5}
			~&  \tabincell{c}{DMPC \\ only} & \tabincell{c}{Hierarchical  \\ CAS}    & \tabincell{c}{DMPC \\ only} & \tabincell{c}{Hierarchical  \\ CAS} \\ \hline
			Test 1 	& 56 &  31 & 169.64 & 306.45 \\
			Test 2& 	63 & 10  & 150.79 & 950.00  \\ 
			Test 3& 	61 & 17  & 155.74  & 558.82 \\ 
			Test 4& 	70 & 18  & 135.71   &527.78\\ 
			Test 5& 	63 & 24  & 150.79   & 395.83 \\ \hline
			Summation & 313 & 100 &  \\\hline
			Mean & ~ & ~ & 152.53 & 547.78 \\\hline\hline
		\end{tabular}
	\begin{tablenotes}
		\footnotesize
		\item{*} CAS: the abbreviation of "collision avoidance system"
	\end{tablenotes}
	\end{center}
\end{table}
Table \ref{tb:probabilistic_sensing_5_UAVs} shows that, the average collision-free distance using the hierarchical collision avoidance system ($547.78m$) is more than three times that of the DMPC-only scheme ($152.53m$). This indicates that the proposed hierarchical strategy is more capable in the uncertain real world.

\section{Conclusion}

In conclusion, this paper studied a three-layered collision avoidance architecture for autonomous multiple fixed-wing UAVs. The effectiveness of the hierarchical collision avoidance system is tested via numerical simulations, in which the result verified the advantage of the proposed methodology in comparison with the DMPC-only collision avoidance scheme. This work is the first attempt of combing several different approaches together to handle complex conflict scenarios of multiple UAVs.

Future work will continue to study the safety management for multiple fixed-wing UAVs.  Firstly, the parameters and algorithms involved in the integrated methodology could to be further optimized to maximize the effect of each layer of the integrated scheme. Secondly, the study on this issue in three-dimensional space is in progress. Besides, physical experiment is also a concern of the authors in future work.

\bibliography{ifacconf}       
\end{document}